\documentclass[pmlr,twoside,tablecaptionbottom]{pgm-jmlr}
\usepackage{pgm-pmlr-dat}

\usepackage{bm}
\firstpageno{1}

\usepackage{colortbl}
\usepackage{pgf, tikz}
\usetikzlibrary{arrows,automata,fit}
\usepgflibrary{shapes.geometric}

\usepackage{subcaption}

\newcommand\independent{\protect\mathpalette{\protect\independenT}{\perp}}
\def\independenT#1#2{\mathrel{\rlap{$#1#2$}\mkern2mu{#1#2}}}

\usepackage[utf8]{inputenc}
\inputencoding{utf8}

\newcommand{\xx}{1}
\newcommand{\yy}{1}

\newcommand{\stage}[2]{\tikz{\node[shape=circle,draw,inner sep=1pt,fill=#1]{$v_{#2}$};}}

\newcommand{\stages}[2]{\tikz{\node[shape=circle,draw,inner sep=1pt,fill=#1,minimum size=0.5cm]{\scriptsize{$v_{#2}$}};}} 

\newcommand{\leaf}{\tikz{\node[shape=circle,draw,inner sep=1.5pt,fill=white] {};}}

\usepackage{booktabs}

\title[Context-Specific Refinements of Bayesian Network Classifiers]{Context-Specific Refinements of Bayesian Network Classifiers}

  \author{
  \Name{Manuele Leonelli} \Email{manuele.leonelli@ie.edu}\\
  \addr School of Science and Technology, IE University, Madrid, Spain
  \AND
  \Name{Gherardo Varando} \Email{gherardo.varando@uv.es}\\
  \addr Image Processing Lab, University of Valencia, Valencia, Spain
  }

\begin{document}
\maketitle

\begin{abstract}
Supervised classification is one of the most ubiquitous tasks in machine learning. Generative classifiers based on Bayesian networks are often used because of their interpretability and competitive accuracy. The widely used naive and TAN classifiers are specific instances of Bayesian network classifiers with a constrained underlying graph. This paper introduces novel classes of generative classifiers extending TAN and other famous types of Bayesian network classifiers. Our approach is based on staged tree models, which extend Bayesian networks by allowing for complex, context-specific patterns of dependence. We formally study the relationship between our novel classes of classifiers and Bayesian networks. We introduce and implement data-driven learning routines for our models and investigate their accuracy in an extensive computational study. The study demonstrates that models embedding asymmetric information can enhance classification accuracy.

\end{abstract}


\section{Introduction}
\label{sec:intro}

We consider the problem of supervised classification of a categorical class variable $C$ given a vector of categorical features $\bm{X}=(X_1,\dots,X_p)$ using generative classifiers. Given a training set of labeled observation $\mathcal{D}=\{(\bm{x}^1,c^1),\dots,(\bm{x}^n,c^n)\}$, a generative classifier aims to learn a joint probability $P(c,\bm{x})$ and assign a non-labeled instance $\bm{x}$ to the most probable a posteriori class found as
\[
\arg\max_{c\in\mathbb{C}}P(c|\bm{x})=\arg\max_{c\in\mathbb{C}}P(c,\bm{x}),
\]
where $\bm{x}^i\in\mathbb{X}$ and $c^i\in\mathbb{C}$. With $\mathbb{X}=\times_{i=1}^{p}\mathbb{X}_i$ and $\mathbb{C}$ we denote the sample spaces of the feature and class variables, respectively.

Bayesian network classifiers (BNCs) \citep{bielza2014discrete,friedman1997bayesian} are the most widely-used class of generative classifiers which factorize $P(c,\bm{x})$ according to a Bayesian network over $\bm{X}$ and $C$. They have been shown to have competitive classification performance with respect to black-box discriminative classifiers while being interpretable and explainable since they explicitly describe the relationship between the features using a simple graph. The famous naive Bayes classifier \citep{minsky1961steps} can be seen as a specific instance of BNCs with a fixed graph structure where no edges between features are allowed.

The main limitation of BNCs is that they can only formally encode symmetric conditional independence. However, there is now a growing amount of evidence that real-world scenarios are better described by more generic, asymmetric types of relationships \citep[e.g.][]{eggeling2019algorithms,leonelli2023context,leonelli2024structural,rios2024scalable}, for instance context-specific ones \citep{boutilier1996context}. There have been limited attempts to extend BNCs to embed asymmetric types of dependence, most notably Bayesian multinets \citep{geiger1996knowledge}, but their use in practice is limited. 

Staged tree classifiers have been recently introduced by \citet{carli2023new}. They have been shown to extend the class of BNCs to embed complex patterns of asymmetric dependence using staged tree models \citep{collazo2018chain,smith2008conditional}. Staged trees are an explainable class of probabilistic graphical models that visually depict dependence using a colored tree. A particular type of staged tree classifier is the \textit{naive} staged tree, which, while having the same complexity as naive Bayes, extends it to account for asymmetric dependences.

In this paper, we introduce novel classes of staged tree classifiers, which can be seen as refinements of  famous sub-classes of BNCs, namely TAN \citep{friedman1997bayesian} and $k$-DB \citep{sahami1996learning} classifiers. We formally investigate the relationship between our novel staged tree classifiers and their BNCs' counterparts. Data-driven learning routines for these novel classes are discussed and implemented. An extensive experimental study compares the classification performance of our novel classifiers to BNCs. The results highlight that these novel classes can increase classification accuracy in some cases by explicitly modelling asymmetric and flexible relationships between features.

\section{Bayesian Network Classifiers}
\label{sec:bn}
\subsection{The Bayesian Network Model}

Let $G=([p],E_G)$ be a directed acyclic graph (DAG) with vertex set $[p]=\{1,\dots,p\}$ and edge set $E_G$. For $A\subset [p]$, we let $\bm{X}_A=(X_i)_{i\in A}$ and $\bm{x}_A=(x_i)_{i\in A}$ where $\bm{x}_A\in\mathbb{X}_A=\times_{i\in A}\mathbb{X}_i$. We say that $P$ is Markov to $G$ if, for $\bm{x}\in\mathbb{X}$, 
\begin{equation}
\label{eq:markov}
P(\bm{x})=\prod_{k\in[p]}P(x_k | \bm{x}_{\Pi_k}),
\end{equation}
where $\Pi_k$ is the parent set of $k$ in $G$. Henceforth, we assume the existence
of a linear ordering $\sigma$ of $[p]$ for which only pairs $(i,j)$ where $i$ appears before $j$ in the order can be in the edge set.

The ordered Markov condition implies conditional independences of the form
\begin{equation}
\label{ci}
X_i \independent \bm{X}_{[i-1]}\,|\, \bm{X}_{\Pi_i}.
\end{equation}
Let $G$ be a DAG and $P$ Markov to $G$. The \emph{Bayesian network} model (associated to $G$) is 
\[
\mathcal{M}_G = \{P\in\Delta_{|\mathbb{X}|-1}\,|\, P \mbox{ is Markov to } G\}.
\]
where $\Delta_{|\mathbb{X}|-1}$ is the ($|\mathbb{X}|-1$)-dimensional  probability simplex.

Let $\mathcal{G}$ be the set of DAGs with vertex set $[p]$ and ordering $\sigma$. We define the space of Bayesian network models over $\bm{X}$ as $\mathcal{M}_{\mathcal{G}}=\cup_{G\in\mathcal{G}}\mathcal{M}_G$

\subsection{Classes of Bayesian Network Classifiers}

BNCs are BNs with vertices $\bm{X}$ and $C$. Although any Bayesian network could be, in principle, used for classification, most commonly, the space of considered DAGs is restricted to those where $C$ has no parents and there is an edge from $C$ to $X_i$ for every $i\in [p]$ \citep[this class is sometimes referred to as Bayesian network-augmented naive Bayes,][]{friedman1997bayesian}. By BNCs, we henceforth refer to such classifiers.

\begin{figure}
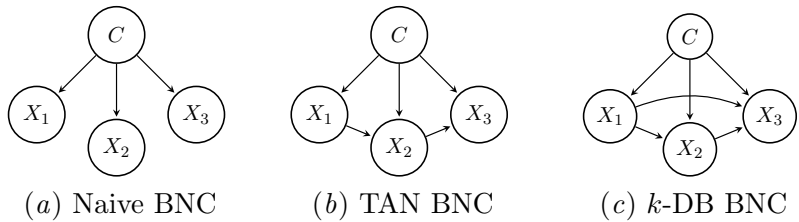

\floatconts
  {fig:bncs}
  {\caption{Examples of BNCs with three features and one class.}}
  {%
    \subfigure[Naive BNC]{\label{fig:naivebnc}%
      \includeteximage[scale = 0.75]{teximage}}%
    \qquad
    \subfigure[TAN BNC]{\label{fig:tanbnc}%
      \includeteximage[scale = 0.75]{tan}}
                \qquad
    \subfigure[$k$-DB BNC]{\label{fig:kdepbnc}%
      \includeteximage[scale = 0.75]{kdep}}
  }
\end{figure}

Subclasses of BNCs entertaining specific properties in the underlying DAG have been defined. The simplest possible model is the so-called naive Bayes classifier \citep{minsky1961steps}, which assumes that the features are conditionally independent, given the class (\figureref{fig:naivebnc}). BNCs of increasing complexity can then be defined by adding dependencies between the feature variables. 
Another commonly used classifier is the TAN BNC \citep{friedman1997bayesian}, for which each feature has at most two parents: the class and possibly another feature (\figureref{fig:tanbnc}). The more generic $k$-DB BNCs \citep{sahami1996learning} assume that each feature can have at most $k$ feature parents (\figureref{fig:kdepbnc}). Naive and TAN classifiers are $k$-DB BNCs for $k=0$ and $k=1$, respectively.

Although BNCs of any complexity can be learned and used in practice, empirical evidence demonstrates that model complexity does not necessarily imply better classification accuracy \citep{bielza2014discrete}. Despite their simplicity, naive and TAN BNCs have been shown to lead to good accuracy in classification problems.

\section{Staged Tree Classifiers}

\subsection{The Staged Tree Model}

Consider a $p$-dimensional random vector $\bm{X}$ taking values in the product sample space $\mathbb{X}$. Let $(V,E)$ be a directed, finite, rooted tree with vertex set $V$, root node $v_0$, and edge set $E$. 
For each $v\in V$, 
let $E(v)=\{(v,w)\in E\}$ be the set of edges emanating
from $v$ and $\mathcal{L}$ be a set of labels. 

An $\bm{X}$-compatible staged tree 
is a triple $(V,E,\theta)$, where $(V,E)$ is a rooted directed tree and:
\begin{enumerate}
    \item $V = {v_0} \cup \bigcup_{i \in [p]} \mathbb{X}_{[i]}$;
		\item For all $v,w\in V$,
$(v,w)\in E$ if and only if $w=\bm{x}_{[i]}\in\mathbb{X}_{[i]}$ and 
			$v = \bm{x}_{[i-1]}$, or $v=v_0$ and $w=x_1$ for some
$x_1\in\mathbb{X}_1$;
\item $\theta:E\rightarrow \mathcal{L}^*=\mathcal{L}\times \cup_{i\in[p]}\mathbb{X}_i$ is a labeling of the edges such that $\theta(v,\bm{x}_{[i]}) = (\kappa(v), x_i)$ for some 
			function $\kappa: V \to \mathcal{L}$. The function $k$ is called the coloring of the staged tree $T$.
\end{enumerate}
	If $\theta(E(v)) = \theta(E(w))$ then $v$ and $w$ are said to be in the same 	\emph{stage}. Therefore, the equivalence classes induced by  $\theta(E(v))$
form a partition of the internal vertices of the tree in \emph{stages}.

Points 1 and 2 above construct  a rooted tree where each root-to-leaf path, or equivalently each leaf, is associated with an element of the sample space $\mathbb{X}$.  Then a labeling of the edges of such a tree is defined where labels are pairs with one element from a set $\mathcal{L}$ and the other from the sample space $\mathbb{X}_i$ of the corresponding variable $X_i$ in the tree. By construction, $\bm X$-compatible staged trees are such that two vertices can be in the same stage if and only if they correspond to the same sample space.

\begin{figure}
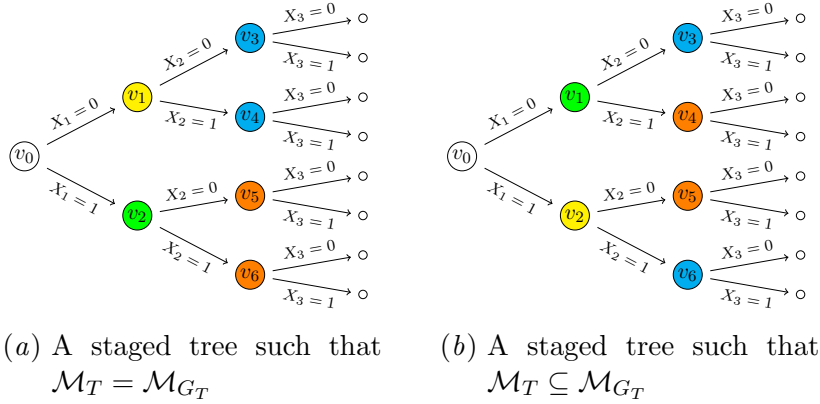

\floatconts
  {fig:sts}
  {\caption{Examples of X-compatible staged trees}}
  {%
    \subfigure[A staged tree such that $\mathcal{M}_T=\mathcal{M}_{G_T}$]{\label{fig:compst}%
      \includeteximage[scale = 0.75]{comp}}%
    \qquad
\subfigure[A staged tree such that $\mathcal{M}_T\subseteq\mathcal{M}_{G_T}$]{\label{fig:xorst}%
      \includeteximage[scale = 0.75]{xor}}
  }
\end{figure}

\figureref{fig:compst} reports an $(X_1,X_2,X_3)$-compatible staged tree over three binary variables. The \textit{coloring} given by the function $\kappa$ is shown in the vertices and
each edge $(\cdot , (x_1, \ldots, x_{i}))$ is labeled with $X_i = x_{i}$. 
The edge labeling $\theta$ can be read from the graph by combining the text label and the 
color of the emanating vertex. 
The staging of the staged tree in \figureref{fig:compst} is given by the partition $\{v_0\}$, $\{v_1\}$, $\{v_2\}$, $\{v_3,v_4\}$ and $\{v_5,v_6\}$.

The parameter space associated to an $\bm X$-compatible staged tree $T = (V, E, \theta)$ 
with 
labeling $\theta:E\rightarrow \mathcal{L}^{*}$ 
is defined as
\begin{equation}
\label{eq:parameter}
	\Theta_T=\Big\{\bm{y}\in\mathbb{R}^{|\theta(E)|} \;|\; \forall ~ e\in E, y_{\theta(e)}\in (0,1)\textnormal{ and }\sum_{e\in E(v)}y_{\theta(e)}=1\Big\}
\end{equation}
Equation~(\ref{eq:parameter}) defines a class of probability mass functions 
over the edges emanating from any internal vertex coinciding with conditional distributions  $P(x_i | \bm{x}_{[i-1]})$, $\bm{x}\in\mathbb{X}$ and $i\in[p]$. In the staged tree in \figureref{fig:compst} the staging $\{v_3,v_4\}$ implies that the conditional distribution of $X_3$ given $X_1=0$, and $X_2 = 0$, represented by the edges emanating from $v_3$, is equal to the conditional distribution of $X_3$ given $X_1=0$ and $X_2=1$. A similar interpretation holds for the staging $\{v_5,v_6\}$. This in turn implies that  $X_3\independent X_2|X_1$, thus illustrating that the staging of a tree is associated with conditional independence statements.

Let $\bm{l}_{T}$ denote the leaves of a staged tree $T$. Given a vertex $v\in V$, there is a unique path in $T$ from the root $v_0$ to $v$, denoted as $\lambda(v)$. The number of edges in $\lambda(v)$ is called  the distance of $v$, and the set of vertices at distance $k$ is denoted by $V_k$.  For any path $\lambda$ in $T$, let $E(\lambda)=\{e\in E: e\in \lambda\}$ denote the set of edges in the path $\lambda$.

The \emph{staged tree model} $\mathcal{M}_{T,\theta}$ associated to the $\bm X$-compatible staged 
	tree $(V,E,\theta)$ is the image of the map
\begin{equation}
\label{eq:model}
\begin{aligned}
\phi_T & : \Theta_T\rightarrow \Delta_{|\bm{l}_T|-1}^{\circ};\\
\phi_T & : y \mapsto p_l = \Big(\prod_{e\in E(\lambda(l))}y_{\theta(e)}\Big)_{l\in \bm{l}_T}
\end{aligned}
\end{equation}
Therefore, staged tree models are such that atomic probabilities are equal to the product of the edge labels in root-to-leaf paths and coincide with the usual factorization of mass functions via recursive conditioning.

Let $\Theta$ be the set of functions $\theta$ from $E$ to $\mathcal{L}^{*}$, that is all possible partitions, or staging, of the staged tree. We define $\mathcal{M}_T=\cup_{\theta\in\Theta}\mathcal{M}_{T,\theta}$. So as $\mathcal{M}_{\mathcal{G}}$ is the union of all possible BN models given a specific ordering, $\mathcal{M}_T$ is the union of all possible staged tree models, that is of all possible stagings, given a specific ordering of the variables.

\subsection{Staged Trees and Bayesian Networks}

Although the relationship between Bayesian networks and staged trees was already formalized 
by \citet{smith2008conditional}, a formal procedure to represent a Bayesian network as a staged tree has been only recently introduced  \citep[e.g.][]{varando2024staged}.  Assume  $\bm{X}$ is topologically ordered with respect to a DAG
$G$ and consider an $\bm X$-compatible staged tree with vertex set $V$,  
edge set $E$ and labeling $\theta$ defined via the 
coloring $\kappa(\bm{x}_{[i]} ) = \bm{x}_{\Pi_{i}}$ of the vertices. The staged tree $T_G$, with vertex set $V$, edge set $E$ and labeling $\theta$
so constructed, is called \emph{the staged tree model of $G$}. 
Importantly,
$\mathcal{M}_G= \mathcal{M}_{T_G}$, i.e. the two models are exactly the same,
since they entail exactly the same factorization of the joint
probability \citep{smith2008conditional}. Clearly, the staging of $T_G$  represents the
Markov conditions associated to the graph $G$.

\citet{varando2024staged} approached the reverse problem of transforming a staged tree into a Bayesian network. Of course, since staged trees represent more general asymmetric conditional independences, given a staged tree $T$ most often there is no Bayesian network with DAG $G_T$ such that $\mathcal{M}_T=\mathcal{M}_{G_T}$. However,  \citet{varando2024staged} introduced an algorithm that, given an $\bm{X}$-compatible  staged tree $T$, finds the minimal DAG $G_T$ such that $\mathcal{M}_T\subseteq \mathcal{M}_{G_T}$. Minimal means that such a DAG $G_T$ embeds all symmetric conditional independences that are in $\mathcal{M}_T$ and that there are no DAGs with less edges than $G_T$ embedding the same conditional independences.

As an illustration, the staged tree in \figureref{fig:compst} can be constructed as the $T_G$ from the Bayesian network with DAG $X_2\leftarrow X_1 \rightarrow X_3$, embedding the conditional independence $X_3\independent X_2 \;|\;X_1$. Conversely, consider the staged tree $T$ in \figureref{fig:xorst}. Such a staged tree does not embed any symmetric conditional independence, only non-symmetric ones, and therefore there is no DAG $G_T$ such that $\mathcal{M}_{G_T}=\mathcal{M}_T$. Furthermore, the minimal DAG $G_T$ such that $\mathcal{M}_T\subseteq\mathcal{M}_{G_T}$ is the complete one since the staging of the tree implies direct dependence between every pair of variables.

\citet{leonelli2022highly} introduced a subclass of staged trees based on the topology of the associated minimal DAG, which will be relevant for the definition of the novel classifiers below. A staged tree $T$ is said to be in the class of $k$-parents staged trees if the maximum in-degree in $G_T$ is less or equal to $k$. For instance, the staged tree in \figureref{fig:compst} is in the class of 1-parent staged trees, whilst the one in \figureref{fig:xorst} is not, since its associated minimal DAG is such that $X_3$ has two parents.

\subsection{Staged Trees for Classification}

\begin{figure}
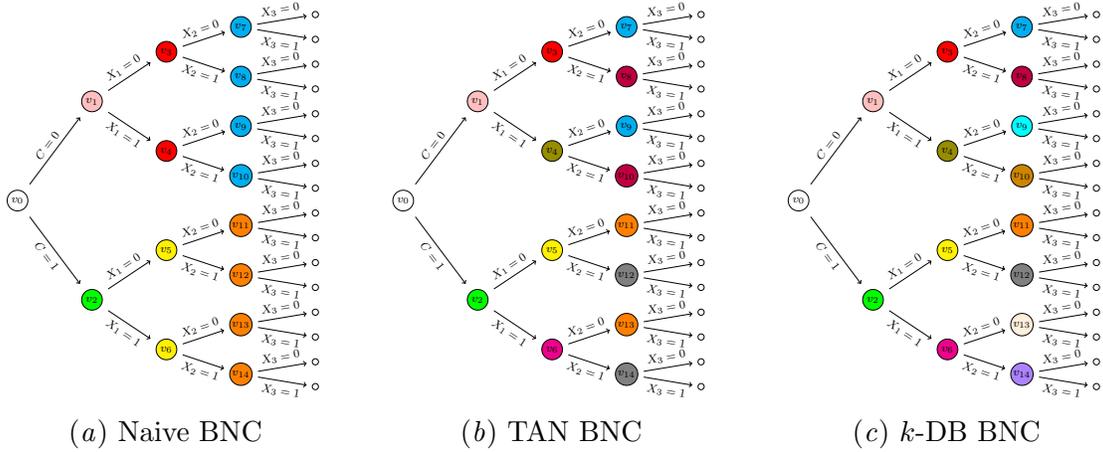

\floatconts
  {fig:sts1}
  {\caption{Staged tree representation of the BNCs in \figureref{fig:bncs}.}}
  {%
    \subfigure[Naive BNC]{\label{fig:naivest}%
      \includeteximage[scale = 0.55]{stnaive}}%
    \qquad
    \subfigure[TAN BNC]{\label{fig:tanst}%
      \includeteximage[scale = 0.55]{sttan}}
                \qquad
    \subfigure[$k$-DB BNC]{\label{fig:kdepst}%
      \includeteximage[scale = 0.55]{kdepst}}
  }
\end{figure}

\citet{carli2023new} discussed how staged tree models can be used for classification purposes. As in \sectionref{sec:intro}, suppose $\bm{X}$ is a vector of features and $C$ is the class variable. A \textbf{staged tree classifier} for the class $C$ and features $\bm{X}$ is a $(C,\bm{X})$-compatible staged tree. The requirement of $C$ being the root of the tree follows from the idea that in most BNCs the class has no parents, so to maximize the information provided by the features for classification. 

All BNCs reviewed in \sectionref{sec:bn} can therefore be represented as staged tree classifiers, by constructing the equivalent $T_G$.  \figureref{fig:sts1} shows the staged trees equivalent to the BNCs in \figureref{fig:bncs}. However, \citet{carli2023new} demonstrated that the class of staged tree classifiers is much larger than that of BNCs. Formally, letting $\mathcal{M}_{\mathcal{G}}^C$ be the space of BNCs and $\mathcal{M}_{T}^C$ the space of $(C,\bm{X})$-compatible staged trees, then $\mathcal{M}_{\mathcal{G}}^C\subset \mathcal{M}_{T}^C$.

\begin{figure}
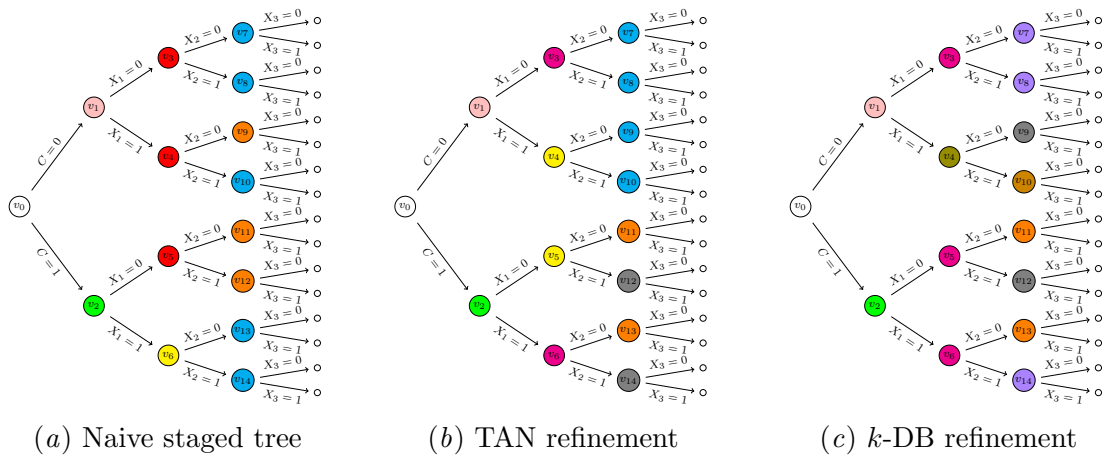

\floatconts
  {fig:stcl}
  {\caption{Examples of subclasses of staged tree classifiers.}}
  {%
    \subfigure[Naive staged tree]{\label{fig:naive}%
      \includeteximage[scale = 0.55]{naive1}}%
    \qquad
    \subfigure[TAN refinement]{\label{fig:tan}%
      \includeteximage[scale = 0.55]{tan1}}
                \qquad
    \subfigure[$k$-DB refinement]{\label{fig:kdep}%
      \includeteximage[scale = 0.55]{kdep1}}
  }
\end{figure}

Since the class of staged tree classifiers is extremely rich, \citet{carli2023new} introduced a subclass of staged tree classifiers termed \textit{naive}. Let $V_k$ be the set of nodes of a tree at distance $k$ from the root. Formally, a $(C,\bm{X})$-compatible staged tree classifiers such that for every $k\leq p$, the set $V_k$ is partitioned into $|\mathbb{C}|$ stages is called naive. The name naive comes from these classifiers having $\sum_{j=1}^p|\mathbb{C}|(|\mathbb{X}_j|-1)+|\mathbb{C}|-1$ free parameters that need to be learned, the same number as for the standard naive Bayes model. An example of a naive staged tree is given in \figureref{fig:naive}. Notice that unlike naive BNCs, which have a fixed DAG structure, the coloring of the vertices must also be learned for data for naive staged trees. \citet{carli2023new} proposed using k-means and hierarchical clustering algorithms for this task.

Just as for generic staged tree classifiers, naive staged trees generalize naive BNCs. Letting $\mathcal{M}_{\mathcal{G}}^{\textnormal{naive}}$ and $\mathcal{M}_T^{\textnormal{naive}}$ be the space of naive BNCs and naive staged tree classifiers, respectively, we have that $\mathcal{M}_{\mathcal{G}}^{\textnormal{naive}} \subset \mathcal{M}_T^{\textnormal{naive}}$. Importantly, \citet{carli2023new} showed via simulation experiments that naive staged tree classifiers can correctly classify parity functions (or 2-XORs), which cannot be captured by naive Bayes models \citep{varando2015decision}.

\section{Context-Specific Classifiers as Refinements of BNCs}
\label{sec:novelsevt}

Next, we introduce novel subclasses of staged tree classifiers that are different from naive staged tree classifiers. These subclasses are inspired by subclasses of BNCs, and they are refined to embed asymmetric patterns of dependence.

\begin{definition}
    A staged tree classifier $T$ is said to be a TAN (or generally $k$-DB) refinement 
	if its minimal DAG $G_T$ is a TAN (or $k$-DB) BNC.
\end{definition}

Examples of TAN and $k$-DB staged tree classifiers are given in \figureref{fig:stcl}. The coloring of the tree is much more flexible than for BNCs, embedding asymmetric patterns of dependence. It can be shown that for these staged tree classifiers $T$ (except for the naive staged tree classifier) their associated $G_T$ are the ones in \figureref{fig:bncs}. 

The following simple result links our new classes of classifiers to $k$-parents staged trees. Although straightforward, this result guides the learning algorithms for the new classes of classifiers since routines already established for $k$-parents staged trees can be simply adapted  to our classifiers.

\begin{proposition}
   If a staged tree classifier is a $k$-DB refinement, then it is in the class of $(k+1)$-parents staged trees. 
\end{proposition}

Notice that naive staged tree classifiers are not necessarily 1-parent staged trees
since they are defined differently from $k$-DB refinements. 
The latter are defined through the minimal DAG, while naive staged trees are defined 
based on the number of parameters of the model. 
The naive staged tree in \figureref{fig:stcl} is such that its minimal DAG is  complete and different to the DAG of a naive Bayes model.
Conversely to naive and generic staged tree classifiers, our novel classes of classifiers are refinements of their DAG equivalent. Let $\mathcal{M}_{\mathcal{G}}^k$ and $\mathcal{M}_T^k$ be the space of $k$-DB BNCs and staged tree classifiers.
\begin{proposition}
$\mathcal{M}_{\mathcal{T}}^k\subseteq\mathcal{M}_G^k$.
\end{proposition}
An analogous result can be stated for TAN classifiers. This means that our classifiers may represent a subset of the decision rules that their BNC counterparts can. However, having a smaller number of parameters that can thus be better estimated from data, our refinements may provide better performance than BNCs when they represent the true data-generating system.

\section{Learning Staged Tree Classifiers}
Learning the structure of a staged tree from data is challenging due to
the fast increase in the size of the tree with the number of the considered random
variables. 
The first learning algorithm used for this purpose was the agglomerative hierarchical clustering (AHC) proposed by \citet{freeman2011bayesian}.
This starts from a staged tree where each vertex is in its own stage and joins the two stages at each iteration leading to the highest increase in a model score,
until no improvement is found. Initially, AHC considered only a Bayesian score, but 
the \texttt{stagedtrees} implementation contains a \emph{backward hill-climbing} method, which is similar
in the optimization technique, and allows using arbitrary scores such as BIC and AIC. 
Since then, other learning algorithms have been proposed,
most often by considering some restricted space of
staged trees \citep[e.g.][]{leonelli2024learning,leonelli2024robust,rios2024scalable},
just as in this paper. 
The novel learning algorithms we introduce follow three steps: 
(i) an initial, appropriate BNC is learned from data; 
(ii) the DAG associated with the learned BNC is transformed into its equivalent staged tree; 
(iii) the AHC algorithm is run starting from the equivalent staged tree from (ii).
We give details of these phases and their implementation in the following sections. 



\subsection{Step (i): Learning a BNC}

To learn BNCs we employ the routines implemented in the 
\texttt{bnclassify} R package~\citep{bnclassify}.
In particular, we consider the following algorithms and \texttt{bnclassify} corresponding implementations:
\begin{itemize}
    \item TAN BNCs obtained by either optimizing the log-likelihood with the Chow-Liu algorithm~\citep[\texttt{tan\_cl,}][]{chowliu} or maximising the cross-validated estimated accuracy (\texttt{tan\_hc}).
    \item k-DB BNCs obtained by greedy optimization of the cross-validated estimated
accuracy (\texttt{kdb}).
\end{itemize}





\subsection{Step (ii): Transforming a DAG into a Staged Tree}

\citet{varando2024staged} defined a conversion algorithm to transform any Bayesian network $G$ into its equivalent staged tree $T_G$.
In our routines we use the implementation provided in the \texttt{stagedtrees}
R package~\citep{carli2022r} by the functions \texttt{as\_sevt} and \texttt{sevt\_fit}.
Notice that by construction the resulting staged tree is a $(k+1)$-parents staged tree.

\subsection{Step (iii): Refining the Staged Tree}

Starting from the staged tree obtained at the previous step, we run the AHC algorithm which only joins stages together (no splitting of stages). We use the \texttt{stages\_bhc} function from the \texttt{stagedtrees} package based on the minimization of the model BIC~\citep{gorgen2022curved}.
This step learns asymmetric dependencies between the features that were joined by an edge in step (i) without adding any new dependence between the features and the class.
Therefore the resulting staged tree classifier is in the appropriate $k$-DB class.

For parameter estimation (i.e. the conditional probabilities of the stages) we use the state-of-the-art maximum likelihood estimation method, possibly with a smoothing parameter to avoid zero probabilities which could harm classification for unseen instances~\citep{bielza2014discrete}.  Under the assumption of the correct stage structures, this method corresponds to the Bayes classifier rule, which minimizes the probability of misclassification~\citep{devroye2013probabilistic}.

\section{Experiments}

We compare  different instantiations of the 
novel paradigm for staged tree classifiers in computational experiments. Specifically, we consider staged tree classifiers refined from 
TAN BNCs (\texttt{sevt\_tan\_cl} and \texttt{sevt\_tan\_hc}) and  k-DB (for $k=3,5$; \texttt{sevt\_3db} and 
\texttt{sevt\_5db}).
We compare their performance to the corresponding BNC classifiers (\texttt{bnc\_tan\_cl}, \texttt{bnc\_tan\_hc}, \texttt{bnc\_3db}, and \texttt{bnc\_5db}). We also consider the naive staged tree classifier (\texttt{sevt\_kmeans\_cmi}) obtained with the k-means clustering of probabilities as proposed by \citet{carli2023new} and implemented in the \texttt{stages\_kmeans} function of \texttt{stagedtrees}.

\subsection{Benchmarks}

\begin{figure}
    \centering
	\includegraphics[width = 0.95\textwidth]{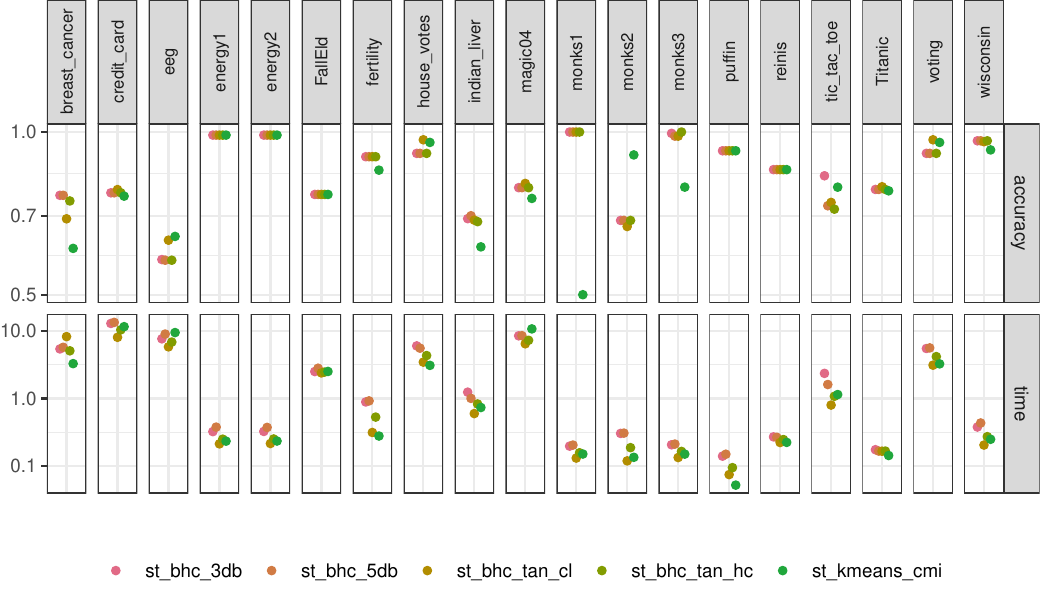}
    \caption{Accuracy and elapsed time for all the staged tree classifiers considered.}
    \label{fig:results}
\end{figure}

\begin{figure}
    \centering
	\includegraphics[width = 0.95\textwidth]{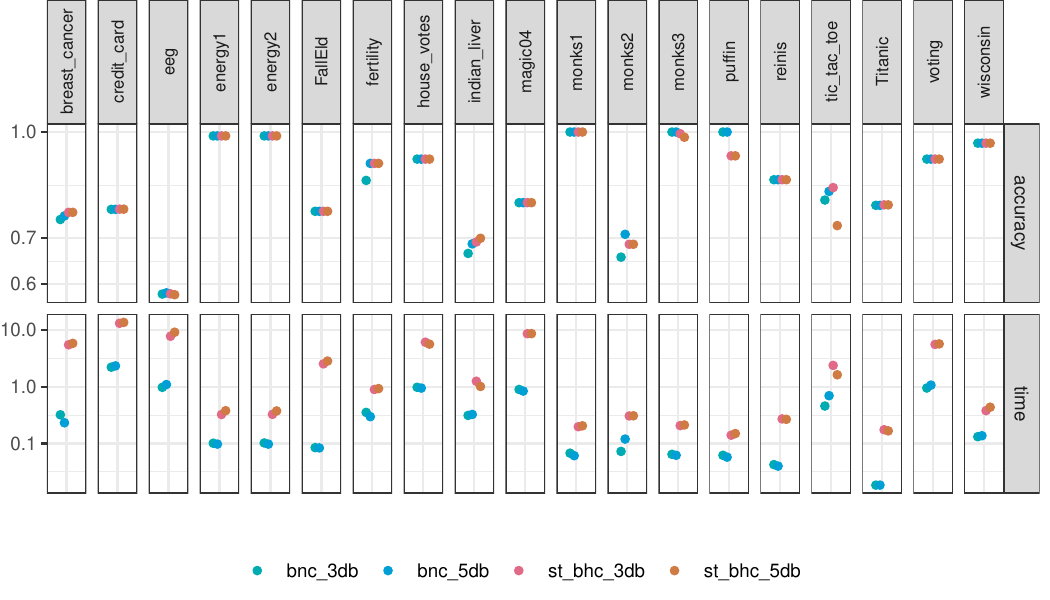}
    \caption{Comparison of median accuracies and elapsed time between $k$-DB BNCs and $k$-DB staged tree classifiers ($k=3,5$).}
    \label{fig:resultsKDB}
\end{figure}

\begin{figure}
    \centering
	\includegraphics[width = 0.95\textwidth]{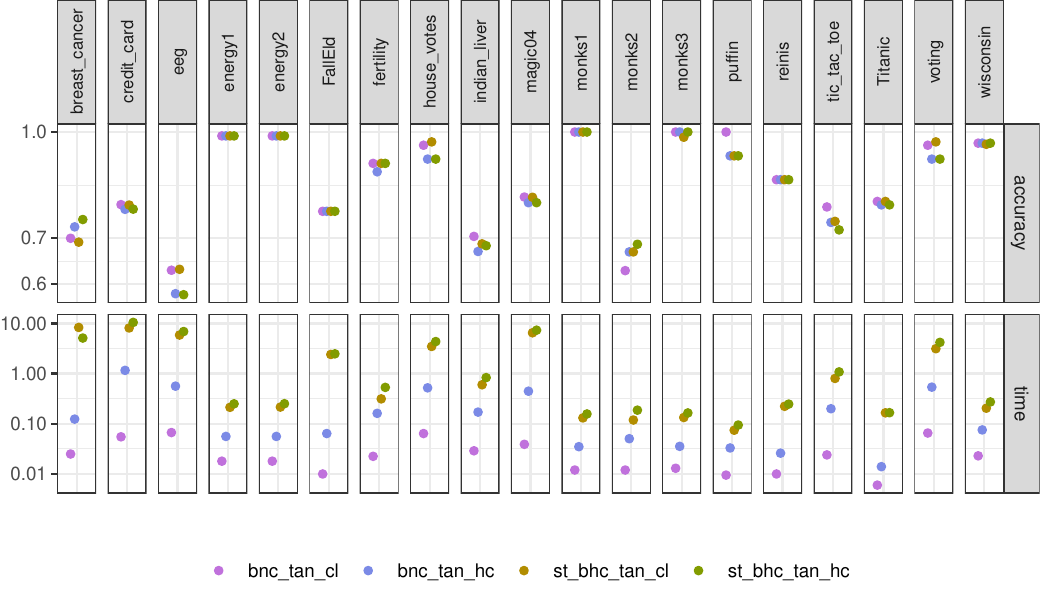}
    \caption{Comparison of median accuracies and elapsed time between TAN BNCs and TAN staged tree classifiers.}
    \label{fig:resultsTAN}
\end{figure}

We compare the considered classifiers across various benchmark datasets, similarly to \citet{carli2023new} but also including some additional datasets with a larger number of features. Details about the datasets are given in Table \ref{table:data}, which also includes the normalized entropy of the class variable as a measure of dataset imbalance. For each dataset, we repeat 10 times an 80\% - 20\% train-test split and we report (Figures~\ref{fig:results}, \ref{fig:resultsKDB}, \ref{fig:resultsTAN}) median accuracies and elapsed times. Median F1 scores, balanced accuracies and precisions are reported in  Appendix~\ref{app:results}.

In Figure~\ref{fig:results} we compare all considered staged tree classifiers. While TAN and $k$-DB classifiers have comparable accuracy across all datasets, the naive staged tree is more variable, in some cases outperforming the others and in others having lower accuracy. All staged tree classifiers require a similar training time. In Figure~\ref{fig:resultsKDB} we compare staged trees $k$-DB classifiers with their corresponding $k$-DB BNCs and in Figure~\ref{fig:resultsTAN} we similarly compare TAN staged trees with corresponding TAN BNCs. It can be observed that overall the accuracy of all approaches is similar. In some instances BNCs outperform staged tree refinements, while in others the reverse can be observed. Considering TAN classifiers only, the \texttt{st\_bhc\_tan\_cl} often outperform the others. As expected, staged tree refinements require more training time, but can still be learned in short time frames.

\subsection{Simulations}

We perform a simulation study to 
explore how the classifiers behave under different scenarios. We consider three different data-generating processes:
\begin{itemize}
    \item \texttt{random\_sevt}: generates a random staged tree model over binary variables, with the ordering $C, X_1, X_2, \ldots, X_p$ and then samples from its induced joint distribution. 
    Random staged trees are generated with the \texttt{random\_sevt} function in the \texttt{stagedtrees} package. The function randomly joins stages starting from the full staged tree and assigns random conditional probabilities uniformly from the probability simplex.

    \item \texttt{linear}: samples independent binary predictors  as $\text{Bernulli}(q)$ where $q \sim \text{Unif}([0,1])$. The class variable $C$ is then obtained as 
    $C =  \operatorname{sign}\left( \sum_{i=1}^p \alpha_i X_i   + \gamma + \varepsilon \right)$, where $\alpha_i, \gamma \sim \text{Unif}([-p,p])$ and $\varepsilon \sim\text{Unif}([-1.2, 1.2])$.

    \item \texttt{xor}: similar to the \texttt{linear} case, but with 
    $C=\operatorname{sign}\left( \prod_{i=1}^p X_i + \varepsilon \right)$.
\end{itemize}

\begin{figure}
    \centering
	\includegraphics[scale=0.8]{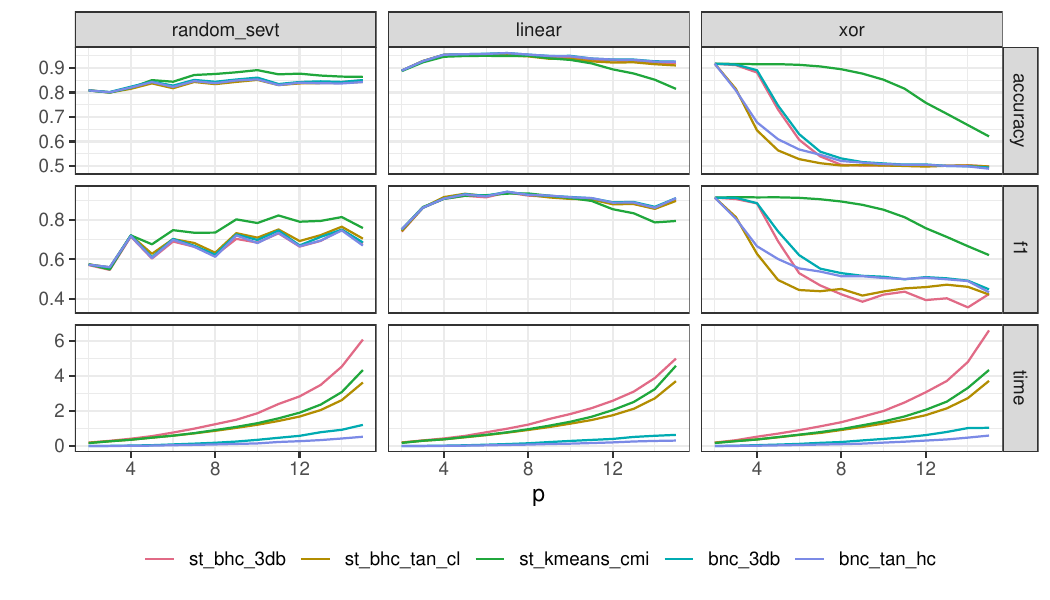}
    \caption{Comparison of median accuracy, F1 and elapsed time in the simulation study.}
    \label{fig:resultsSIM}
\end{figure}

We vary the number of predictors ($p$) ranging from 2 to 15.
For each case, we generate $100$ repetitions of $1000$ training and $1000$ test samples generated with the above three processes. 
In Figure~\ref{fig:resultsSIM} we report median accuracies, F1 scores and elapsed times across repetitions. The naive staged tree outperforms all other approaches for all values of $p$ under the \texttt{random\_sevt} and \texttt{xor} simulation scenarios, while it has a decrease of performance in the \texttt{linear} data-generating scenario for large values of $p$. The other algorithms have comparable performance except for the \texttt{xor} data generating scenario.
The results in the \texttt{xor} scenario are partially to be expected: TAN and kDB BNC, as well as the staged tree refinements, cannot represent complex functions~\citep{varando2015decision}, and thus, as the number of predictors increases, their performance decrease drastically. 
As partially anticipated in \citet{carli2023new}, the naive staged tree classifiers, here implemented with k-means clustering, can estimate optimal decision functions across a diverse range of data-generating processes, while only slightly worsening the sample-efficiency for larger $p$.

\section{Discussion}

The paper introduced novel learning routines for subclasses of staged tree classifiers, which enhance BNCs with context-specific patterns of dependence. Experimental studies showed that in some scenarios they can outperform BNCs, although the difference in accuracy appears to be only marginal.

As with BNCs, different and specific 
parameter estimation techniques could be envisioned, such as discriminative learning~\citep[e.g.][]{pernkopf2011maximum} and weighted schemes~\citep[e.g.][]{frank2002locally}.  Furthermore, adjusted class prior probabilities could be used to address imbalanced datasets, when different classification errors entail distinct costs~\citep{wong2021multinomial}.

The simulation study suggests that naive staged tree classifiers are highly expressive, with high accuracy in the \texttt{xor} data generating scenario, where all other classifiers perform poorly. This fact led us to believe that naive staged trees may be able to theoretically represent any decision rule, or at least a set of decision rules much larger than those expressible by subclasses of BNCs. A formalization of this statement and its associated proof is currently being developed.

\bibliography{references}

\appendix

\newpage 

\section{Benchmark Datasets Details}

\begin{table}[htpb]
\floatconts
  {table:data}
  {\caption{Details about the 19 datasets included in the experimental study.}}
  {
  \scalebox{0.95}{
\begin{tabular}{ccccc}
\toprule
Dataset & \# observations & \# variables & \# atomic events & imbalance measure \\
\midrule
\texttt{breast\_cancer} &277&10&332640 &  0.872\\
\texttt{credit\_card} &30000 & 12 & 18432 & 0.762 \\
\texttt{eeg} & 14979 & 15 &  32768 &   0.992\\
\texttt{energy1} &768&9&1728 & 1.000\\
\texttt{energy2} &768&9&1728 & 1.000 \\
\texttt{fallEld} &5000&4&64 & 0.888 \\
\texttt{fertility} &100&10&15552 & 0.529 \\
\texttt{house\_votes} & 232 & 17 & 131072 &0.997 \\
\texttt{indian\_liver} & 579 & 11 & 15552& 0.862 \\
\texttt{magic04} & 19020 & 11 & 118098 & 0.936 \\
\texttt{monks1} &432&7&864 & 1.000 \\
\texttt{monks2}&432&7&864 & 0.914 \\
\texttt{monks3} &432&7&864 & 0.998 \\
\texttt{puffin} &69&6&768 & 0.999 \\
\texttt{reinis} & 1841 & 6 & 64 &0.587\\
\texttt{ticTacToe} &958&10&39366 & 0.931\\
\texttt{titanic} &2201&4&32 & 0.908 \\
\texttt{voting} &435&17&131072 & 0.997 \\
\texttt{wisconsin} &683&10&1024 & 0.934 \\
\bottomrule
\end{tabular} }}
\end{table}

\newpage 

\section{Additional Results}
\label{app:results}
\begin{figure}[htpb]
    \centering
	\includegraphics[width = 0.9\textwidth]{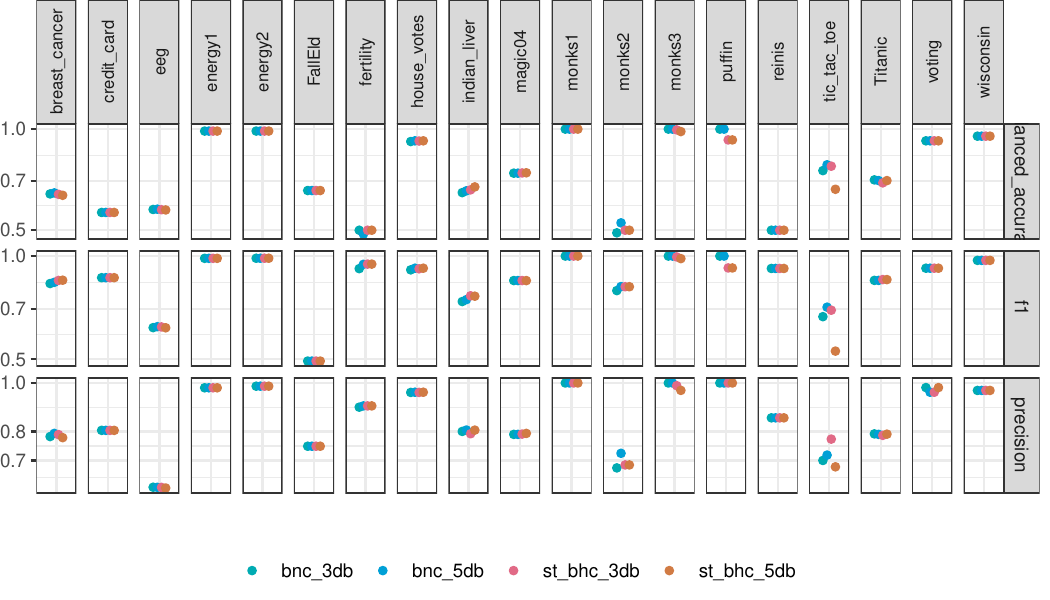}
    \caption{Median F1 scores, balanced accuracies and precisions for $k$-DB BN and staged tree classifiers.}
    \label{fig:f1KDB}
\end{figure}

\begin{figure}[htpb]
    \centering
	\includegraphics[width = 0.9\textwidth]{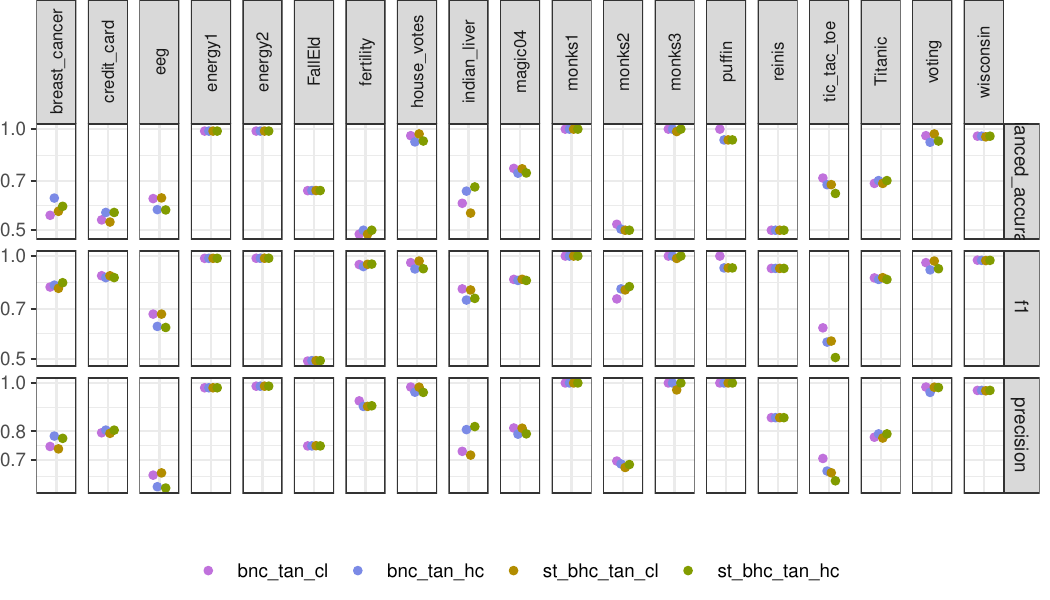}
    \caption{Median F1 scores, balanced accuracies and precisions for TAN BN and staged tree classifiers.}
    \label{fig:f1TAN}
\end{figure}

\end{document}